\documentclass[10pt,twocolumn,letterpaper]{article}

\usepackage[pagenumbers]{cvpr}

\usepackage[dvipsnames]{xcolor}
\usepackage[normalem]{ulem}
\usepackage{lipsum}
\usepackage{multirow}
\usepackage[skip=1 pt]{caption} 

\usepackage{tikz}
\usepackage{wrapfig}
\usepackage{amsmath,bm}
\usepackage{relsize}
\usetikzlibrary{positioning,calc}

\definecolor{cvprblue}{rgb}{0.21,0.49,0.74}
\usepackage[pagebackref,breaklinks,colorlinks,allcolors=cvprblue,hypertexnames=false]{hyperref}

\definecolor{darkgreen}{RGB}{30,150,30}
\definecolor{darkblue}{RGB}{0,0,127}
\definecolor{darkyellow}{RGB}{171,133,0}
\definecolor{darkred}{RGB}{180,20,20}
\definecolor{darkmagenta}{RGB}{127,0,127}
\definecolor{darkcyan}{RGB}{0,127,127}
\definecolor{chromeyellow}{rgb}{1.0, 0.65, 0.0}
\definecolor{amber}{rgb}{1.0, 0.75, 0.0}

\newcommand{\methodName}{L4P\xspace}
\newcommand{\plucker}{Pl{\"u}cker\xspace}

\title{L4P: Towards Unified \uline{L}ow-Level \uline{4}D Vision \uline{P}erception}

\newcommand{\mychar}{
  \begingroup\normalfont\centering
  \includegraphics[width=0.12\textwidth]{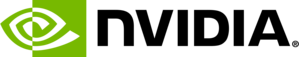}
  \endgroup
}

\author{
  Abhishek Badki$^*$ \quad  Hang Su$^*$ \quad Bowen Wen \quad Orazio Gallo \\ \vspace{-0.3em} \\
  \mychar{} \\
  \small\url{https://research.nvidia.com/labs/lpr/l4p}\\
}

\begin{document}

\maketitle

\newcommand\blfootnote[1]{
  \begingroup
  \renewcommand\thefootnote{}\footnote{#1}
  \addtocounter{footnote}{-1}
  \endgroup
}
\blfootnote{\hspace{-1em} $^*$ indicates equal contribution.}

%! TEX Root = ../main.tex
\begin{abstract}
The spatio-temporal relationship between the pixels of a video carries critical information for low-level 4D perception tasks.
A \emph{single} model that reasons about it should be able to solve \emph{several} such tasks well.
Yet, most state-of-the-art methods rely on architectures specialized for the task at hand.
We present \emph{L4P}, a feedforward, general-purpose architecture that solves low-level 4D perception tasks in a unified framework.
L4P leverages a pre-trained ViT-based video encoder and combines it with per-task heads that are lightweight and therefore do not require extensive training.
Despite its general and feedforward formulation, our method is competitive with existing specialized methods on both dense tasks, such as depth or optical flow estimation, and sparse tasks, such as 2D/3D tracking.
Moreover, it solves all tasks at once in a time comparable to that of single-task methods.
\vspace{-1em}
\end{abstract}
    
\section{Introduction}\label{sec:intro}
%! TEX Root = ../main.tex

Large collections of videos are our most complete and compact source of priors about the world.
Much like text did for large-language models, the corpus of videos we amassed over the years allowed video-language models (VLMs) to produce remarkable zero-shot results on high-level vision tasks such as video captioning, video question answering, and others.
However, zero-shot, low-level 3D and 4D vision perception tasks, such as depth from video, tracking, optical flow, and others remain a challenge.
Pre-trained diffusion models fine-tuned on target-domain data showed potential on dense vision perception tasks (\eg, depth~\cite{ke2024repurposing,hu2024depthcrafter}, flow~\cite{saxena2024surprising}, \etc), but the fine-tuning makes them task-specific, and therefore limits their ability to leverage priors across multiple tasks at once.
Sparse vision perception tasks, such as tracking, are even more challenging to tackle with a general foundation model, because their representation does not fit naturally into pixel-aligned data structures, 
and are typically addressed with optimization-based approaches~\cite{wang2023omnimotion} or specialized architectures~\cite{karaev2023cotracker,doersch2023tapir}.

Can we leverage the priors learned from a large body of video data and solve \emph{multiple} low-level 4D vision perception tasks, both \emph{dense} and \emph{sparse}, with a \emph{unified} architecture with strong generalization abilities?

\begin{figure}
    \captionsetup{skip=5pt}
    \centering
    \includegraphics[width=\columnwidth]{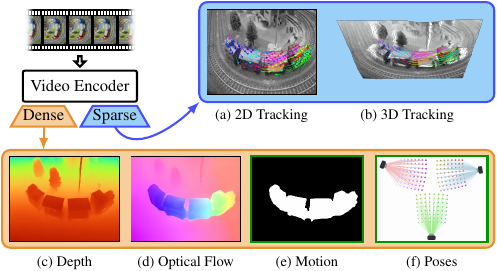}
    \caption{We propose L4P, a general-purpose architecture that solves several low-level 4D perception tasks. Building on a pre-trained video encoder and lightweight per-task heads, our unified model is competitive with existing methods specialized to solve individual tasks. L4P can easily be extended to additional tasks, (e) and (f), without compromising performance.}
    \label{fig:teaser}
    \vspace{-1em}
\end{figure}

\begin{figure}
    \captionsetup{skip=3pt}
    \centering
    \includegraphics[width=\columnwidth]{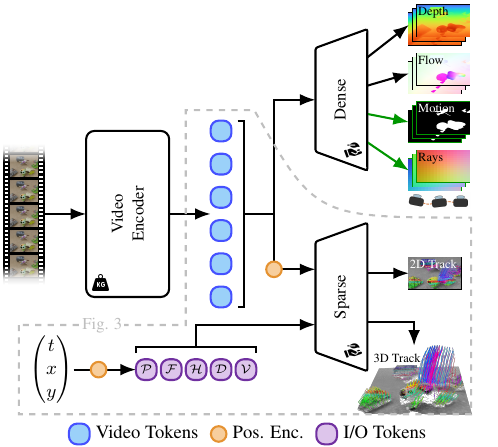}
    \caption{
        We use a pre-trained video encoder as tokenizer, and combine it with per-task lightweight heads.
        For sparse tasks we define additional query tokens: the point to track, $\mathcal{P}$, the corresponding feature token, $\mathcal{F}$, and output tokens (heatmap $\mathcal{H}$, depth $\mathcal{D}$, and visibility $\mathcal{V}$).
    }\label{fig:method}
\end{figure}

This is challenging: we need a shared backbone that is strong and versatile. 
On one hand, it must be general enough to allow for pre-training on auxiliary tasks. 
On the other hand, it must be able to support diverse tasks that require fundamentally different representations, such as dense 2D planes and 3D tracks.
We tackle this challenge by combining a pre-trained video masked auto-encoder (VideoMAE)~\cite{tong2022videomae,wang2023videomaev2} with per-task, lightweight heads (Figure~\ref{fig:method}).
VideoMAEs offer a feedforward, online mechanism to tokenize videos within a small computational envelope.
Indeed they have been successfully employed for a variety of mid- and high-level vision tasks~\cite{wang2023videomaev2}, but their ability to capture the spatio-temporal relationship between pixels is underexplored in the context of low-level 4D perception.
For dense tasks, we couple VideoMAE with heads based on the dense prediction transformer (DPT), which has been shown to perform well on depth estimation, image segmentation, and others~\cite{ranftl2021dpt}.
For sparse tasks, we focus on tracking.
We posit that tracking is important for perception because understanding fine-grained, complex motions and the physical interaction between objects is critical to downstream applications, including 3D reconstruction~\cite{lei2024mosca, wang2023vggsfm} and robotics manipulation~\cite{xu2024flow,wen2022you,wen2024anypointtrajectory, bharadhwaj2024track2act}.
We formulate the problem of tracking as that of predicting \emph{2D} heatmaps for queried pixels with associated depth and visibility tokens. 
This formulation allows us to tackle sparse and dense tasks within a \emph{unified} model.
To implement it, we augment a general-purpose head with a novel memory mechanism to track points for arbitrarily long videos.
We finetune our model for multiple tasks at once on a small collection of synthetic datasets.
Yet, thanks to the training curriculum we propose, and to VideoMAE's priors, it shows strong generalization to real data.

Our formulation presents several desirable properties and advantages.
First, the pre-trained VideoMAE model allows us to tap into priors learned from large datasets that are potentially different and more varied than those typically used for low-level 4D perception.
It also affords us efficient computation: our system solves all tasks in roughly 300ms for a 16-frame video chunk ($\sim$19ms/frame), which is comparable to, or faster than, methods specialized for each task (see Table 1 in the Supplementary).
Moreover, combining it with per-task heads allows us to train a relatively small number of parameters for new tasks, which we show by freezing the system and adding a head for motion-based segmentation and camera pose estimation (marked in green in Figures~\ref{fig:teaser}~and~\ref{fig:method}).
Lastly, but perhaps most importantly, breaking the architecture into a general VideoMAE and per-task heads offers a mechanism to solve both dense and sparse tasks with a \emph{unified} model (Figure~\ref{fig:teaser}).
We emphasize that our main target is a general, multi-task framework, rather than the performance on any specific tasks. 
Nonetheless, our model performs competitively on the individual tasks, often on par with state-of-the-art methods.
This is remarkable because the baselines we compare against are task-specific, carefully designed for, and specialized to excel at their respective tasks.
Finally, VideoMAEs are already used as encoders for VLMs~\cite{wang2024internvideo2,2023videochat}, and we speculate that training them to reason about low-level 4D perception may impart those capabilities to the downstream VLMs they may be used with, though validating this is outside the scope of our paper.
In summary:

\begin{itemize}
    \item We show that priors from video representation learning can be leveraged for low-level 4D perception tasks with strong generalization capabilities.
    \item We present a feedforward video models that adopts a general-purpose architecture for multiple 4D perception tasks jointly---including both dense and sparse ones---and demonstrate its effectiveness on in-the-wild video inputs.
    \item We combine the general-purpose architecture with a novel memory mechanism to support long-range tracking to tackle the limited context of the video encoder.
    \item Our model demonstrates strong performance, solving multiple tasks jointly with results on par with state-of-the-art, within timeframes typical of single-task approaches. 
\end{itemize}

\section{Related Works}\label{sec:related}
%! TEX Root = ../main.tex
Our method unites the strong generalization capabilities of pre-trained foundation models with lightweight task-specific heads. 
In this section, we review relevant literature for both the foundation models and the individual tasks. 

\subsection{Foundation models for vision perception}
Self-supervised pre-training of large models on huge unlabeled data has shown great success.
Among them, Vision Transformers (ViT)~\cite{vit} pre-trained with masked autoencoding (MAE)~\cite{he2022masked} have become a common choice to fine-tune for many vision tasks, Segment Anything (SAM)~\cite{kirillov2023SAM} being a notable example.
Several approaches have been proposed for spatio-temporal representation learning for videos: VideoMAE~\cite{tong2022videomae,wang2023videomaev2}, MAE-ST~\cite{feichtenhofer2022maest}, V-JEPA~\cite{assran2025vjepa2}, etc.
We adopt VideoMAEv2~\cite{wang2023videomaev2}, which introduces a dual masking strategy that allows them to efficiently scale up the model to a billion parameters and to effectively leverage priors from large data.
Their representation is effective for action recognition, and more recently is combined with language models~\cite{wang2024internvideo2,2023videochat}.
We show how to leverage them for low-level 4D perception tasks and achieve results competitive with specialized SOTA methods.

\subsection{Dense Prediction Tasks}
\noindent
\textbf{Depth estimation.}
Stereo depth estimation approaches~\cite{furukawa2009accurate, galliani2015massively,schonberger2016pixelwise, yao2018mvsnet,huang2018deepmvs, sayed2022simplerecon} and single image depth estimation approaches~\cite{eigen2014depth,ranftl2020midas,ranftl2021dpt, yang2024depthanything,yang2024depthanythingv2} have made significant progress.
We focus on estimating depth from videos of dynamic scenes, for which several solutions exist~\cite{luo2020consistent,kopf2021robust,teed2018deepv2d, wang2023nvds}.
Pioneered by DUSt3R~\cite{wang2024dust3r}, an emerging paradign estimates pointmaps instead of depthmaps. 
Several recent approaches~\cite{zhang2024monst3r, leroy2024mast3r,wang2025cut3r,wang2025vggt} follow this paradigm for joint depth and camera pose estimation.
To tackle the data scarcity problem, the state-of-the-art approaches leverage priors from foundation models: DepthAnything~\cite{yang2024depthanything,yang2024depthanythingv2} uses DINOv2~\cite{oquab2023dinov2}, DUSt3R~\cite{wang2024dust3r} uses cross-view completion pre-training~\cite{weinzaepfel2023croco}, and Marigold~\cite{ke2024repurposing} and DepthCrafter~\cite{hu2024depthcrafter} use image and video diffusion priors respectively.
We benefit from large-scale pre-training by using a video encoder pre-trained with MAE.

\noindent
\textbf{Optical flow estimation.}
Though straightforward when posed as a dense task, optical flow traditionally requires specialized architectures~\cite{teed2020raft,sun2018pwc,wang2024gflow,xu2023unifying}.
Most related to ours are approaches that take multi-frame inputs~\cite{ren2019fusion,janai2018unsupervised,shi2023videoflow,dong2024memflow}.
Similar to depth estimation, recent works adopt priors from large-scale pre-training, \eg via diffusion priors~\cite{saxena2024surprising} or cross-view completion pre-training~\cite{weinzaepfel2023croco}.

\noindent
\textbf{Motion-based segmentation.}
While early learning-based approaches rely on combining appearance features with flow~\cite{bharadhwaj2024track2act,fragkiadaki2015motionseg,jain2017fusionseg,Tokmakov2017motionv2,Tokmakov2017motionv1}, more recent works~\cite{bideau2019moanet, yang2021rigidmotion} extract geometric properties from flow before using it to train a classification network.
However, they are affected by noisy flow and are limited by their two-frame formulation.

\noindent
\textbf{Camera pose estimation.}
Classical structure-from-motion solutions~\cite{schoenberger2016sfm, snavely2006phototour} work robustly and accurately in constrained settings, while learning-based solutions~\cite{tartanvo2020corl,teed2021droid,li2025megasam,wang2024dust3r,wang2024spann3r,zhang2024monst3r,wang2025cut3r,wang2025vggt} address challenging scenarios like limited image observations, dynamic and texture-less regions, etc.
We cast camera pose estimation as a dense task by representing cameras as dense bundles of rays~\cite{grossberg2001genralimagemodel, zhang2024raydiffusion} using 6-D \plucker coordinates~\cite{plucker1828} and estimating them using our dense head.

We treat motion-based segmentation and camera pose estimation as additional tasks, by leveraging the priors from our video model trained on depth, flow and tracking tasks, and only training task-specific heads.
Thus, we maintain main tasks performance while supporting additional tasks.

\subsection{Sparse Prediction Tasks}
Tracking Any Point (TAP) in a video has many applications~\cite{wen2024anypointtrajectory, bharadhwaj2024track2act,lei2024mosca, wang2023vggsfm,cheng2023segment}.
Particle Video~\cite{harley2022particle}, PIPs~\cite{bian2023contextpips} and TAP-Net~\cite{doersch2022tap} lay the initial foundation by adopting ideas from optical flow approaches.
On the other hand, OmniMotion~\cite{wang2023omnimotion} optimizes a volumetric representation for each video to solve this task and is time-consuming.
TAPIR~\cite{doersch2023tapir} introduces the idea of coarse-to-fine track estimation, BootsTAPIR~\cite{doersch2024bootstap} further improves it by adopting a self-supervised learning, and CoTracker~\cite{karaev2023cotracker} proposes to jointly track multiple points to leverage spatial correlations.
SpaTracker~\cite{spatracker} is a feedforward method for 3D point tracking that lifts pixels to 3D using input depth maps and applies CoTracker's tracking formulation in 3D. 
Concurrent works, TAPIP3D~\cite{tapip3d} and SpatialTrackerV2~\cite{xiao2025spatialtracker}, adopt a similar approach but use posed RGBD inputs; notably, SpatialTrackerV2 jointly optimizes point tracks and initial camera poses via iterative bundle adjustment, which is inefficient for real-time settings.
Seurat~\cite{cho2025seurat} shows that dense 2D tracks alone contain enough information to be lifted to 3D. 
In contrast to these prior works, which often rely on specialized architectures and additional inputs such as depth maps or 2D point tracks, we focus on efficiently unifying 2D and 3D point tracking with other tasks, paving the way for general purpose low-level 4D perception for real-time scenarios.
\section{Method}\label{sec:method}
%! TEX Root = ../main.tex
\begin{figure*}[htbp]
    \captionsetup{skip=4pt}
    \centering
    \includegraphics[width=0.8\textwidth]{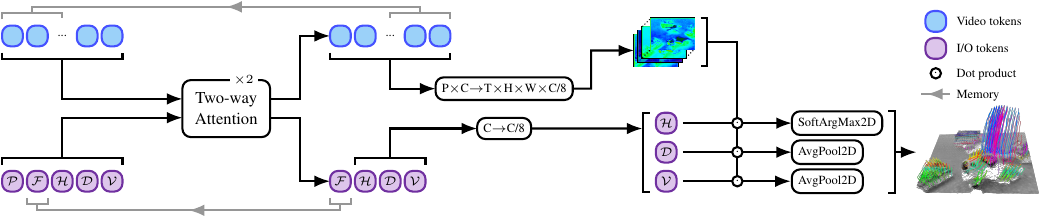}
    \caption{
    Sparse head. The video and I/O tokens (query $\mathcal{P}$oint, query point $\mathcal{F}$eature, and query point $\mathcal{H}$eat map, $\mathcal{D}$epth, and $\mathcal{V}$isibility) are processed by a two-way attention layer. The reshaped and resized per-frame feature maps and processed output tokens are combined via a dot product. We also introduce a memory mechanism to combine video and the $\mathcal{F}$eature tokens from time $t$ and $t+1$.
    }
    \label{fig:sparse_head}
\end{figure*}

We provide an overview of our approach in Figure~\ref{fig:method}.
Our model uses a pre-trained ViT-based video encoder~\cite{wang2023videomaev2} (Section~\ref{sec:method_mae}) to capture spatio-temporal features in an RGB video clip of length $T$.
We use lightweight task-specific heads that decode the video features for low-level 3D/4D perception tasks.
For pixel-wise dense tasks like depth, flow, motion-based segmentation, and camera rays estimation (for camera pose estimation), we propose an extension of the DPT architecture~\cite{ranftl2021dpt} that allows us to use them for videos, instead of just images, as in their existing formulation (Section~\ref{sec:method_dense}).
For the sparse task of tracking any pixel in a video, we take inspiration from the head architecture proposed in SAM~\cite{kirillov2023SAM}~(Section~\ref{sec:method_sparse}).
Given a pixel queried in any frame of the input video, we extend the head, also originally designed to work for images, to decode video tokens into a 2D trajectory, its depth with respect to the camera, and its visibility in each frame~(Section~\ref{sec:windowTrack}).
VideoMAE extracts video tokens from temporal windows of fixed length $T$, and cannot process longer sequences.
We propose a memory mechanism to track points in arbitrarily long videos~(Section~\ref{sec:memory}).

\subsection{Video Masked Auto-Encoders}\label{sec:method_mae}

Motivated by its scalable and powerful pre-training and architecture, we use the ViT-based video encoder from VideoMAEv2~\cite{wang2023videomaev2}.
The encoder works with videos of size $T \times H \times W$ and uses a spatio-temporal patch size of $t \times h \times w$. It uses cube embedding~\cite{tong2022videomae} to transform an input video into a sequence of tokens, 
which are then processed by the spatio-temporal attention blocks to generate video tokens $\mathcal{S} \in \mathbb{R}^{P \times C}$, where $P$ is the number of tokens and $C$ is the embedding dimension.
We run the video encoder only once per video clip, and then apply the lightweight heads to decode the tokens to the desired outputs.
For the point tracking task, we can independently prompt these tokens to track many points in parallel. 

\subsection{Dense Prediction Heads}\label{sec:method_dense}
Dense prediction tasks produce outputs with spatial dimensions aligned with their inputs, typically at the same resolution $H \times W$.  
A wide array of common computer vision problems can be formulated as dense prediction tasks. 
In this work, we explore depth estimation, optical flow estimation, motion-based segmentation, and camera pose estimation as examples.

We adopt the DPT~\cite{ranftl2021dpt} architecture for dense prediction due to its proven performance and efficiency on single-image depth estimation. 
DPT progressively assembles and combines tokens from various layers, thus capturing both local and global spatial structures.
We adapt DPT to work with the 3D tokens from VideoMAE by mapping the video tokens to 3D feature maps and enabling temporal reasoning by introducing 3D convolutions inside the DPT head.
We find that this modification is enough to bring in temporal consistency with minimal computation overhead.
For most of our dense tasks, the DPT heads differ only in the final layer, which outputs one channel for depth and motion-based segmentation, and two channels for optical flow.
We cast camera pose estimation as a dense task by estimating a bundle of rays for each camera, and recover camera parameters from the estimated rays~\cite{zhang2024raydiffusion} (see Supplementary for more details).

For videos longer than $T$ frames, we run inference with stride $T/2$ and enforce consistency in the overlapping frames between temporal windows.
For depth, we align the predictions with an affine transformation.
This strategy has no effect on the individual windows for relative depth, but can greatly improve long-term temporal consistency.
For optical flow and motion-based segmentation, we simply overwrite the predictions in the overlap between windows.
For pose estimation, we align camera trajectories across overlapping windows using 3D similarity transformations between their uplifted point clouds.

\subsection{Sparse Prediction Heads}\label{sec:method_sparse}

Given a video and pixel prompt, $(t_i,x_i,y_i)$, we want to estimate the corresponding 3D trajectory, $\mathcal{T}_i=\{\hat{x}_i(t),\hat{y}_i(t),\hat{d}_i(t),\hat{v}_i(t)\}_{t=0}^{S-1}$, where at time $t$, $(\hat{x}_i(t),\hat{y}_i(t))$ denotes the 2D track location, $\hat{d}_i(t)$ is the track depth with respect to camera, and $\hat{v}_i(t)$ is the track visibility indicating if a track is visible or occluded at time $t$.
This is a challenging task since it requires tracking the pixel in 2D when visible, tracking it through occlusions, and reasoning about its depth along the track.
Moreover, our video encoder has limited temporal context, since it can only process videos with fixed temporal window of $T$ frames, and we want to enable tracking for arbitrarily long videos with $S>T$ frames.
This makes adapting a general-purpose head particularly challenging.
To tackle this, we introduce an online approach.
We design a head that allows us to estimate the 3D track for an input pixel prompt within the temporal context ($T$ frames) of the video encoder.
For online estimation beyond $T$ frames, we propose a memory mechanism for our head and a recipe to train it efficiently (see Figure~\ref{fig:sparse_head}).

\subsubsection{Tracking within the Temporal Context}\label{sec:windowTrack}
Posing the sparse tracking task within our unified framework requires special care. 
Instead of directly estimating point-track positions, we propose to represent tracks using dense probability heatmaps.
Casting tracking as a problem of estimating pixel-aligned 2D maps affords us a consistent representation between sparse and dense tasks, which is critical for using a shared backbone.
To achieve this, we build on the prompt-encoding and mask-decoding mechanisms from SAM~\cite{kirillov2023SAM}.
We encode the input pixel prompt using 3D positional encoding and a learnable embedding to generate input point token $\mathcal{P}$ with embedding dimension of $C$.
Similarly, we define output tokens with learnable embeddings to estimate different components of a 3D track: a heatmap token ($\mathcal{H}$) to estimate the 2D pixel position of the track across the video, a depth token ($\mathcal{D}$), and a visibility token ($\mathcal{V}$). 
Input and output tokens interact with the video tokens, $\mathcal{S}$, also encoded with 3D positional encoding, using a two-way attention mechanism to decode the video features.
These video features are then reshaped and upsampled, and a final inner product with the processed output tokens gives us output masks of size $T \times H \times W$.
For 2D track estimation, we interpret this output mask as a probability heatmap that encodes the 2D track position and apply a 2D soft-argmax operation to estimate the 2D track position $(\hat{x}_i(t),\hat{y}_i(t))$ at each frame $t$.
For depth and visibility, we simply apply a 2D average pooling operation, followed by exponential and sigmoid operations respectively to estimate the track depth $\hat{d}_i(t)$ and the visibility $\hat{v}_i(t)$ at each frame $t$.
This design also allows us to query points anywhere in the video and track them in parallel.
We use two lightweight, two-way attention layers~\cite{kirillov2023SAM}.
We use 3D convolutions to enable temporal reasoning.

\subsubsection{A Memory Mechanism for Long Videos}\label{sec:memory} 
To track beyond a single window of length $T$ frames, we propose an online strategy.
A na\"{i}ve approach would be to chain tracks across windows.
Given two consecutive and overlapping windows, and a 2D track estimated in the first one, we can use a point on the track in the temporal overlap between the two windows as the query for tracking in the second one.
To pick a good point to chain the tracks, we can select the one with highest visibility score.
However, this solution is brittle for two reasons.
First, a tracked point may not be visible in the overlap between the windows.
To tackle this problem, inspired by Karaev~\etal~\cite{karaev2023cotracker}, we introduce a track-feature token $\mathcal{F}$ that is passed to subsequent windows as an additional prompt (see Figure~\ref{fig:sparse_head}).
However, unlike Karaev~\etal, we do not initialize it explicitly with the local appearance around the query point, so the two-way attention head is free to capture the most useful information to track through occlusions.
Second, the na\"{i}ve solution described above does not allow the system to reason across temporal windows, which makes it prone to drifting or to losing tracks.
The track-feature tokens help, but to provide even more cross-window information, we pass the video tokens decoded by the two-way attention stage of the current window to the next, as shown in Figure~\ref{fig:sparse_head}.
We achieve this by projecting the decoded video tokens in the overlapping region via a linear layer, and by adding them to the corresponding incoming video tokens to the two-way attention stage in the next window.
Our memory strategy based on these two mechanisms is critical to allow proper reasoning across temporal windows as shown by the comparison in ablation study in Section~\ref{sec:ablation}.

\noindent\textbf{Online training.} Training the memory mechanism requires unrolled-window training~\cite{karaev2023cotracker}, in which we compute the video features for all the overlapping windows in a video of length $S$, and then compute the tracks for the entire video in an online fashion.
However, training such an approach end-to-end is prohibitive due to memory constraints.
To alleviate this, we adapt a multi-stage training strategy.
First, we train only for a single window but train all the parameters of our network.
In the next stage, we freeze all but the last few layers of our video encoder, and fine-tune it along with the tracking head for unrolled window training.
In Table~\ref{tab:ablation}, we show this approach improves the performance over the na\"{i}ve chaining approach.
\section{Implementation}\label{sec:impl}
%! TEX Root = ../main.tex

The training curriculum is critical to leverage the VideoMAE priors and fine-tune our system for all tasks at once.

\noindent
\textbf{Training datasets.}
We use the video encoder from Wang~\etal~\cite{wang2023videomaev2}, which is pre-trained on 1.35M video clips for masked auto-encoding.
To fine-tune our model, we use four synthetic datasets covering various types of annotations, and rely on the priors from the pre-trained video encoder for generalization.
We use Kubric~\cite{greff2022kubric}, a synthetic dataset in which multiple objects interact, annotated with ground truth (GT) depth, flow, and 2D/3D point tracking.
To include videos with long 3D trajectories, we add PointOdyssey~\cite{zheng2023pointodyssey} and DynamicReplica~\cite{karaev2023dynamicstereo}.
Both are synthetic datasets with animated characters in mostly indoor scenes.
Both have depth GT and DynamicReplica also offers optical flow GT.
To further increase scene diversity, we also include TartanAir~\cite{tartanair2020iros}, which provides GT for flow and depth.

\noindent\textbf{Architectures.}
Our video encoder processes $16 \times 224\times224$ clips.
It uses a patch size of $2\times14\times14$, which results in $P=2048$ video tokens, and an embedding dimension of $C=1408$.
It has 40 encoder blocks, and we use the output from blocks 14, 21, 28, 36 for DPT heads for dense tasks, while the sparse heads use features only from the last block (block 39).
Feeding the sparse and dense heads with tokens from different blocks is critical for maintaining the performance on dense tasks while we fine-tune our model to train the memory mechanism for the tracking tasks, as we discuss below.
For a $16\times224\times224$ video clip, our method generates the outputs for all our tasks, including the additional tasks, in $\sim$300ms on an NVIDIA A6000 GPU.
This corresponds to $\sim$19ms for a single frame, which is competitive with single-task approaches (see Supplementary for detailed comparisons).
However, our method's latency may prevent its use for applications that require strict real-time performance.

\noindent\textbf{Training.}
Perhaps unsurprisingly, training our system to maximize the performance on all tasks at once requires particular care.
We initialize the video encoder using a pre-trained VideoMAE~\cite{wang2023videomaev2} and fine-tune the complete system in three stages.
All the stages use a batch size of 8 and are trained on a single 8-GPU (NVIDIA A100) node for 100k iterations.
In all the stages, we construct a batch of multiple tracks per video for the tracking task.
In the first stage, we train end-to-end for depth, flow and point tracking tasks on a single window of $T=16$ frames and train only on Kubric.
In the second stage of training, we add the remaining three datasets and further fine-tune the model on a single window for all the tasks.
Since not all the datasets offer GT for all tasks, we form our batches by sampling two videos from each of the four datasets.
This forces each batch to provide training signal for all task, and thus prevents training instability.
In the third stage, we further fine-tune our model for the tracking tasks using unrolled-window training and our memory mechanism for online tracking.
We train on videos of length $S=40$ by using 4 overlapping windows of size 16 frames and a stride of 8.
Due to memory constraints, in the third stage, we freeze all the parameters, except the last three layers (37-39) of the video encoder and the sparse task head.
This allows us to maintain the performance on depth and flow, while training the memory mechanism to improve the tracking tasks.
Training takes around 4 days for all the stages combined.

\noindent\textbf{Losses.}
We use the SILog~\cite{eigen2014depth} loss for depth and L1 loss for optical flow.
For tracking, we use L1 loss for 2D track positions, scale-invariant loss for track depth (similar to dense depth), and binary cross entropy loss for track visibility.
Like with the choices of tasks heads, we pick the most widely used losses for each of our tasks.
However, since we train for multiple tasks at once, weighting the losses appropriately is critical.
We find the loss weights empirically by bringing the losses in the same order of magnitude and then doing a small hyperparameter search around those weights.

Please refer to the Supplementary for additional important implementation details.

\section{Experiments}\label{sec:expts}
%! TEX Root = ../main.tex

For each task included in our main model, we compare against a few recent SOTA methods on standard datasets.
Our focus is a general, multi-task framework rather than achieving SOTA on any one task.
Overall, we demonstrate competitive performance across all tasks, with computation times that are comparable to or faster than those of single-task approaches.
We also show that L4P can be extended to new tasks without degrading existing performance. 

%! TEX Root = ../main.tex
\begin{table}
\begin{center}
    \resizebox{.8\columnwidth}{!}{
    \begin{tabular}{rccccccccccc}
        \toprule
        & \multicolumn{2}{c}{Sintel ($\sim$50 frames)} & \multicolumn{2}{c}{ScanNet} (90 frames) & \multicolumn{2}{c}{KITTI} ($\sim$110 frames) & \multicolumn{2}{c}{Bonn} (110 frames) & \multicolumn{2}{c}{NYUv2} (1 frame)\\
        \cmidrule(lr{0.1em}){2-3}\cmidrule(lr{0.1em}){4-5}\cmidrule(lr{0.1em}){6-7}\cmidrule(lr{0.1em}){8-9}\cmidrule(lr{0.1em}){10-11}
        & AbsRel $\downarrow$ & $\delta_1$ $\uparrow$ & AbsRel $\downarrow$ & $\delta_1$ $\uparrow$ & AbsRel $\downarrow$ & $\delta_1$ $\uparrow$ & AbsRel $\downarrow$ & $\delta_1$ $\uparrow$ & AbsRel $\downarrow$ & $\delta_1$ $\uparrow$  \\
        \midrule
        Marigold~\cite{ke2024repurposing} & $0.532$ & $0.515$ & $0.166$ & $0.769$ & $0.149$ & $0.796$ & $0.091$ & $0.931$ & $0.070$ & $0.946$  \\
        DA~\cite{yang2024depthanything} & $0.325$ & $0.564$ & $0.130$ & $0.838$ & $0.142$ & $0.803$ & $0.078$ & $0.939$ & $\mathbf{0.042}$ & $\mathbf{0.981}$  \\
        DA-V2~\cite{yang2024depthanythingv2} & $0.367$ & $0.554$ & $0.135$ & $0.822$ & $0.140$ & $0.804$ & $0.106$ & $0.921$ & $0.043$ & $0.978$  \\
        \midrule
        NVDS~\cite{wang2023nvds} & $0.408$ & $0.483$ & $0.187$ & $0.677$ & $0.253$ & $0.588$ & $0.167$ & $0.766$ & $0.151$ & $0.780$  \\
        ChronoDepth~\cite{shao2024chronodepth}  & $0.587$ & $0.486$ & $0.159$ & $0.783$ & $0.167$ & $0.759$ & $0.100$ & $0.911$ & $0.073$ & $0.941$  \\
        MonST3R~\cite{zhang2024monst3r} & $0.335$ & $0.585$ & - & - & $0.104$ & $0.895$ & $0.063$ & $0.964$ & $0.091$ & $0.888$  \\ 
        DepthCrafter~\cite{hu2024depthcrafter} & $0.270$ & $0.697$ & $0.123$ & $0.856$ & $0.104$ & $0.896$ & $0.071$ & $0.972$ & $0.072$ & $0.948$  \\ 
        \methodName & $\mathbf{0.219}$ & $\mathbf{0.700}$ & $\mathbf{0.071}$ & $\mathbf{0.953}$ & $\mathbf{0.084}$ & $\mathbf{0.935}$ & $\mathbf{0.056}$ & $\mathbf{0.973}$ & $0.084$ & $0.916$  \\
        \bottomrule
    \end{tabular}%
    }
\end{center}
\caption{\textbf{Depth estimation results.} 
We compare our full model with single-image (Row 1-3) and video depth estimation baselines (Row 4-7).
On video datasets (all except NYUv2), our model consistently performs better than DepthCrafter, the closest competition, and by a large margin on ScanNet and KITTI.
Frames indicate average video length for each dataset.
\vspace{-4mm}
}\label{tab:depth}
\end{table}

\begin{figure}
    \captionsetup{skip=5pt}
    \centering
    \small
    \includegraphics{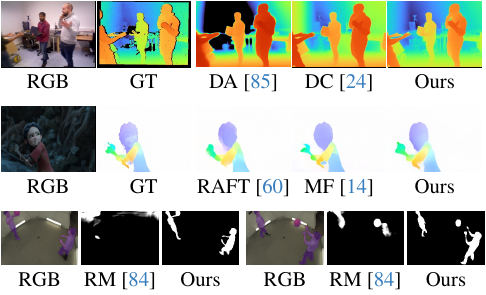}
    \caption{
        Comparisons with SOTA methods. We run inference on videos but show one frame here.
        The purple masks show GT motion segmentation.
        }\label{fig:vis_res}
        \vspace{-7mm}
\end{figure}

\subsection{Video Depth Estimation}

We follow DepthCrafter~\cite{hu2024depthcrafter} and evaluate video depth estimation on a collection of five datasets.
We do not use any of the datasets for training our models or the baselines to better understand their generalization abilities.
When estimating depth for videos longer than $T=16$ frames, we follow the online inference and alignment strategies explained in Section~\ref{sec:method_dense}.
There is an inherent scale-ambiguity in the estimated depthmaps. 
We follow the common practice of aligning linearly the estimation with the GT before calculating evaluation metrics.
The alignment is done for all the frames at once, and is carried out in \emph{disparity} space via least-square fitting.
For comparison on single image datasets, we repeat the single frame 16 times to compute our estimations.
We report two metrics: AbsRel ($\text{mean}(|\hat{\mathbf{d}}-\mathbf{d}| / \mathbf{d}))$ and $\delta_1$ (ratio of pixels satisfying $\max(\mathbf{d}/\hat{\mathbf{d}}, \hat{\mathbf{d}}/\mathbf{d})<1.25$), where $\mathbf{d}$ represents GT, and $\hat{\mathbf{d}}$ is depth estimation after alignment. 
We upsample our estimations from $224\times224$ to each dataset's original resolution for evaluation.
We consider video approaches including NVDS~\cite{wang2023nvds}, ChronoDepth~\cite{shao2024chronodepth}, DepthCrafter~\cite{hu2024depthcrafter}, MonST3R~\cite{zhang2024monst3r}, as well as single-image ones, including Marigold~\cite{ke2024repurposing} and DepthAnything~\cite{yang2024depthanything,yang2024depthanythingv2}. 

Our results show strong performance against both single-image and video depth approaches on the four video datasets (Table~\ref{tab:depth}). 
Since \methodName is a video approach, applying it on single images from NYUv2 does not provide the necessary temporal context for it to perform well. 
DepthCrafter and MonST3R similarly suffer on NYUv2.
Figure~\ref{fig:vis_res} (first row) shows a qualitative comparison. 
\methodName produces a level of details comparable to that of diffusion models such as DepthCrafter, while generally capturing more accurate relative scales.
It is also worth noting that our approach performs competitively on KITTI, despite the fine-tuning not including driving scenarios.

%! TEX Root = ../main.tex

\begin{table}
  \begin{center}
      \resizebox{0.8\columnwidth}{!}{
      \begin{tabular}{rccccccc}
          \toprule
          & \multicolumn{2}{c}{Spring} & \multicolumn{2}{c}{Sintel} & \multicolumn{2}{c}{Virtual KITTI} \\
          \cmidrule(lr{0.1em}){2-3}\cmidrule(lr{0.1em}){4-5}\cmidrule(lr{0.1em}){6-7}
          & $EPE\downarrow$ & $EPE<1\uparrow$ & $EPE\downarrow$ & $EPE<1\uparrow$ & $EPE\downarrow$ & $EPE<1\uparrow$ \\
          \midrule
          RAFT*~\cite{teed2020raft}      &  $0.13$ & $98.4$ & $1.31$ & $85.1$ & $1.13$ & $77.5$ \\
          MemFlow~\cite{dong2024memflow} &  $0.11$ & $\mathbf{98.7}$ & $\mathbf{1.03}$ & $\mathbf{87.2}$ & $0.72$ & $85.4$ \\
          \methodName                    &  $\mathbf{0.09}$ & $\mathbf{98.7}$ & $1.06$ & $84.6$ & $\mathbf{0.63}$ & $\mathbf{86.6}$ \\
          \bottomrule
      \end{tabular}
      }
  \end{center}
  \caption{
        \textbf{Optical flow estimation results.} 
        We compare with SOTA flow methods: RAFT (two-frame, marked with *) and MemFlow (multi-frame).
        We evaluate cross-dataset generalization at 224x224 resolution.
        Our approach is competitive at this resolution, despite using generic architecture.
    }
    \label{tab:flow}
\end{table}

\subsection{Multi-Frame Optical Flow Estimation}

We use the Spring~\cite{mehl2023spring}, Sintel~\cite{butler2012sintel}, and Virtual KITTI~\cite{cabon2020vkitti2} datasets for evaluation.
These video datasets are not used to train ours or other approaches we compare against, allowing us to evaluate generalization ability.
For each dataset, we use all the videos and all the non-overlapping 16-frame clips from each video.
The input frames are resized to $224\times224$ for all evaluation.
We use the Endpoint Error (EPE), as well as a more robust metric, ratio of EPE $<1$, for the evaluation. 

We consider two baselines for comparison. 
RAFT~\cite{teed2020raft}, a competitive and widely used two-frame approach, creates dense pairwise pixel features and uses recurrent updates to estimate optical flow. 
MemFlow~\cite{dong2024memflow}, a recently published work, ranks among the top methods on the Spring benchmark.
It is a multi-frame approach that relies on a memory mechanism to leverage temporal context. 
Quantitatively, \methodName compares favorably to both RAFT and MemFlow (Table~\ref{tab:flow}).
Our model captures both small and large motions and presents more precise motion boundaries (Figure~\ref{fig:vis_res}, second row). 
In addition, multi-frame approaches like MemFlow and ours generally have an edge in temporal stability (see Supplementary). 
Unlike many specialized approaches, our model currently only operates on low-resolution videos and further work is needed to enable efficient high-res estimation.
We discuss this in the Limitations (Section~\ref{sec:limitations}).

%! TEX Root = ../main.tex

\begin{table}
    \begin{center}
        
    \resizebox{.7\columnwidth}{!}{
        \begin{tabular}{rccccccc}
            \toprule
            & Aria & DriveTrack & PStudio & \multicolumn{3}{c}{\textbf{Overall}} \\
            \cmidrule(lr{0.1em}){2-4}\cmidrule(lr{0.1em}){5-7}
            & 2D-AJ $\uparrow$ & 2D-AJ $\uparrow$ & 2D-AJ $\uparrow$
            & 2D-AJ $\uparrow$ & APD $\uparrow$ & OA $\uparrow$ \\
            \midrule
            TAPIR~\cite{doersch2023tapir}         & $48.6$ & $57.2$ & $48.7$ & $53.2$ & $67.4$ & $80.5$ \\
            BootsTAPIR~\cite{doersch2024bootstap} & $54.7$ & $\mathbf{62.9}$ & $\mathbf{52.4}$ & $\mathbf{59.1}$ & $\mathbf{74.7}$  & $85.6$ \\
            CoTracker~\cite{karaev2023cotracker}  & $54.2$ & $59.8$ & $51.0$ & $57.2$ & $74.2$ & $84.5$ \\
            \midrule
            \methodName (2D Only)                 & $\mathbf{55.5}$ & $53.0$ & $48.7$ & $52.4$ & $68.2$  & $\mathbf{88.7}$  \\
            \methodName                           & $52.4$ & $50.0$ & $48.1$ & $50.2$ & $66.1$ & $\mathbf{88.5}$  \\
            \toprule
        \end{tabular}
    }
    \end{center}
    \caption{\textbf{Evaluation of 2D tracking on TAPVid-3D (\textit{full\_eval} split).}
    2D GT trajectories are obtained by projecting 3D GT trajectories onto 2D. 
    Though behind 2D SOTA approaches, \methodName is competitive once fine-tuned only for 2D tracking (``2D Only").
    \vspace{-2mm}
    }
    \label{table:tapvid3d2dtrack}
\end{table}

%! TEX Root = ../main.tex

\begin{table}
    \begin{center}
        \resizebox{\columnwidth}{!}{
        \setlength{\tabcolsep}{3pt}
        \begin{tabular}{rccccccccccccc}
            \toprule
            & \multicolumn{3}{c}{Aria} & \multicolumn{3}{c}{DriveTrack} & \multicolumn{3}{c}{PStudio} & \multicolumn{3}{c}{\textbf{Overall}} \\
            \cmidrule(lr{0.1em}){2-4}\cmidrule(lr{0.1em}){5-7}\cmidrule(lr{0.1em}){8-10}\cmidrule(lr{0.1em}){11-13}
            & 3D-AJ $\uparrow$ & APD $\uparrow$ & OA $\uparrow$ & 3D-AJ $\uparrow$ & APD $\uparrow$ & OA $\uparrow$ & 3D-AJ $\uparrow$ & APD $\uparrow$ & OA $\uparrow$
            & 3D-AJ $\uparrow$ & APD $\uparrow$ & OA $\uparrow$ \\
            \midrule
            Static Baseline       & $4.9$ & $10.2$ & $55.4$ & $3.9$ & $6.5$ & $80.8$ & $5.9$ & $11.5$ & $75.8$ & $4.9$ & $9.4$ & $70.7$ \\
            TAPIR + CM        & $7.1$ & $11.9$ & $72.6$ & $8.9$ & $14.7$ & $80.4$ & $6.1$ & $10.7$ & $75.2$ & $7.4$ & $12.4$ & $76.1$ \\
            CoTracker + CM    & $8.0$ & $12.3$ & $78.6$ & $11.7$ & $\mathbf{19.1}$ & $81.7$ & $8.1$ & $13.5$ & $77.2$ & $9.3$ & $15.0$ & $79.1$ \\
            BootsTAPIR + CM   & $9.1$ & $14.5$ & $78.6$ & $\mathbf{11.8}$ & $18.6$ & $83.8$ & $6.9$ & $11.6$ & $81.8$ & $9.3$ & $14.9$ & $81.4$ \\
            \midrule
            TAPIR + ZD      & $9.0$ & $14.3$ & $79.7$ & $5.2$ & $8.8$ & $81.6$ & $10.7$ & $18.2$ & $78.7$ & $8.3$ & $13.8$ & $80.0$ \\
            CoTracker + ZD  & $10.0$ & $15.9$ & $87.8$ & $5.0$ & $9.1$ & $82.6$ & $11.2$ & $19.4$ & $80.0$ & $8.7$ & $14.8$ & $83.4$ \\
            BootsTAPIR + ZD & $9.9$ & $16.3$ & $86.5$ & $5.4$ & $9.2$ & $85.3$ & $11.3$ & $19.0$ & $82.7$ & $8.8$ & $14.8$ & $84.8$ \\
            TAPIR-3D              & $2.5$ & $4.8$  & $86.0$ & $3.2$ & $5.9$ & $83.3$ & $3.6$ & $7.0$ & $78.9$ & $3.1$ & $5.9$ & $82.8$ \\
            SpatialTracker        & $9.9$ & $16.1$ & $89.0$ & $6.2$ & $11.1$ & $83.7$ & $10.9$ & $19.2$ & $78.6$ & $9.0$ & $15.5$ & $83.7$ \\
            \methodName           & $\mathbf{11.1}$ & $\mathbf{17.7}$ & $\mathbf{90.3}$ & $6.5$ & $11.3$ & $\mathbf{87.8}$ & $\mathbf{18.4}$ & $\mathbf{28.1}$ & $\mathbf{87.5}$ & $\mathbf{12.0}$ & $\mathbf{19.0}$ & $\mathbf{88.5}$ \\ 
            \bottomrule
        \end{tabular}%
        }
    \end{center}
    \caption{
        \textbf{Evaluation of 3D tracking on TAPVid-3D (\textit{full\_eval} split).} 
        The top approaches combine 2D point tracking approaches with COLMAP (CM)~\cite{schoenberger2016sfm}, while the bottom ones, including ours are feedforward.
        ``ZD" refers to ZoeDepth.
        Our approach outperforms previous approaches on average across all the metrics.
        \vspace{-5mm}
        }
    \label{table:tapvid3d3dtrack}
\end{table} 

\begin{figure*}
    \centering
    \includegraphics[width=0.9\textwidth]{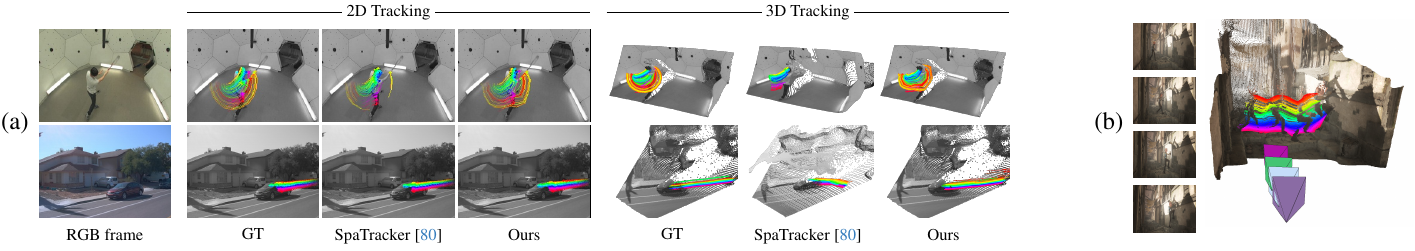}
    \caption{
    Comparisons on 2D/3D tracking (a). L4P can estimate camera poses, depth, and tracking in the same reference system (b).
    }\label{fig:track}
    \vspace{-4mm}
\end{figure*}

\subsection{Sparse 2D/3D Track Estimation}

We evaluate on TAPVid-3D~\cite{koppula2024tapvid3d}, a benchmark containing long-range 3D point trajectories over three datasets: Aria~\cite{pan2023aria}, DriveTrack~\cite{sun2020waymodataset}, and PStudio~\cite{joo2015pstudio}.
It introduced several baselines by combining SOTA 2D point tracking approaches with depth solutions (eg. ZoeDepth~\cite{bhat2023zoedepth}, COLMAP~\cite{schoenberger2016sfm,schoenberger2016mvs}).
We mainly compare against CoTracker~\cite{karaev2023cotracker}, TAPIR~\cite{doersch2023tapir}, and BootsTAPIR~\cite{doersch2024bootstap} for 2D tracking, and SpaTracker~\cite{spatracker} for 3D tracking.

The benchmark evaluates both 3D and 2D tracking approaches, and uses metrics that measure the ability to predict point visibility using an occlusion accuracy metric (OA), the accuracy of predicted trajectories in the visible regions (APD), and joint occlusion and geometric estimation (AJ).
To resolve the scale ambiguity in depth estimation, the benchmark uses global median scaling by computing the median of the depth ratios between the estimated and ground-truth 3D tracks over all the points and frames in a video.
We use the \textit{full\_eval} split evaluation numbers provided in the TAPVid-3D benchmark for the comparison.

On 3D tracking, we outperform SpaTracker (Table~\ref{table:tapvid3d3dtrack}), demonstrating the effectiveness of our general architecture compared to a specialized one.
Approaches that combine 2D track estimation with COLMAP perform better on the DriveTrack~\cite{sun2020waymodataset} dataset.
A reason could be that many query points are on static vehicles, for which COLMAP gives accurate depth.
Such COLMAP-based baselines, however, perform poorly on Aria~\cite{pan2023aria} and PStudio~\cite{joo2015pstudio}, which are mostly dynamic.
We show a comparison with SpaTracker in Figure~\ref{fig:track}.

On 2D tracking, we are slightly behind the SOTA 2D tracking approaches (Table~\ref{table:tapvid3d2dtrack}).
Our approach becomes more competitive when we fine-tune our model only for the 2D tracking task.
We believe our reduced performance on 2D tracking comes from working at lower image resolution, $224\times224$ for us as compared to $384\times512$ for CoTracker, and $256\times256$ for others, and a lack of task-specific tricks, like tracking multiple points together (CoTracker), or assuming access to all frames in the video and performing a global track-refinement (TAPIR and BootsTAPIR), both of which could also benefit our tracking head.

\subsection{Ablations}\label{sec:ablation}
To understand the contribution of different components of our approach, we perform an ablation study for depth, flow and 3D point tracking, as shown in Table~\ref{tab:ablation}.
For each of these tasks, we report average numbers over the datasets not used in our training: for depth we use datasets in Table~\ref{tab:depth}, for optical flow we use the Spring dataset, and for tracking we use the \textit{minival} split from the TAPVid-3D~\cite{koppula2024tapvid3d} benchmark.
Our main contribution is a method that can leverage the priors of a pre-trained VideoMAE for dense and sparse low-level perception tasks.
To show the usefulness of our end-to-end fine-tuning strategy, we compare against a pre-trained and frozen VideoMAE, where we only fine-tune the task-specific heads. 
Table~\ref{tab:ablation} shows that our fine-tuned VideoMAE (row 4) produces better results than the pre-trained and frozen VideoMAE across all tasks (row 2). 
A version trained end-to-end from scratch results in worse performance (row 1), which shows that our system successfully builds on the priors of the pre-trained VideoMAE.
Training our system for all tasks jointly (row 4) performs roughly on par as training for each task individually (row 3), which shows effectiveness of our training strategy to leverage VideoMAE priors for multiple tasks at once.
Finally, by adding the proposed memory mechanism for tracking and using our multi-stage training process for online tracking, we obtain improvements for point tracking, while maintaining the performance on other tasks (row 5 vs. 4).

\begin{table}
    \centering
    \resizebox{.95\columnwidth}{!}{
      \begin{tabular}{l|cccc}
        \multicolumn{1}{c|}{} & \begin{tabular}[c]{@{}c@{}}Depth\\ AbsRel$\downarrow$ / $\delta_1\uparrow$\end{tabular} & \begin{tabular}[c]{@{}c@{}}Optical flow\\ EPE$\downarrow$ / EPE$<1\uparrow$\end{tabular} & \begin{tabular}[c]{@{}c@{}}3D Track\\ 2D-AJ$\uparrow$ / 3D-AJ$\uparrow$\end{tabular} \\ \cline{1-4}
        \multicolumn{1}{l|}{\methodName (from scratch, w/o mem)} & 0.274 / 0.560 & 0.319 / 95.0 & 15.9 / 1.2 \\
        \multicolumn{1}{l|}{\methodName (VideoMAE frozen, w/o mem)} & 0.140 / 0.830 & 0.114 / 98.3 & 33.3 / 4.2 \\
        \multicolumn{1}{l|}{\methodName (task-specific, w/o mem)} & \textit{0.108} / \textit{0.894} & \textbf{0.084} / \textbf{98.8} & 39.4 / 8.4 \\
        \multicolumn{1}{l|}{\methodName (w/o mem)} & \textbf{0.103} / \textbf{0.895} & \textit{0.092} / \textit{98.7} & \textit{40.4} / \textit{8.6} \\
        \multicolumn{1}{l|}{\methodName}           & \textbf{0.103} / \textbf{0.895} & \textit{0.092} / \textit{98.7} & \textbf{49.0} / \textbf{10.9} \\
        
      \end{tabular}
    }
    \caption{\textbf{Ablation study.} 
    See Section~\ref{sec:ablation} for an analysis.}
    \label{tab:ablation}
    \vspace{-4mm}
\end{table}

\subsection{Additional Tasks}

L4P can be extended to new tasks, \emph{without} degrading the performance of the original tasks.
We do this by freezing our video encoder trained for the original tasks, and training a task-specific head.
Here we provide two examples.

\noindent\textbf{Motion-based segmentation.}
We train on the Kubric~\cite{greff2022kubric}, and evaluate on the Virtual KITTI (VKITTI)~\cite{cabon2020vkitti2} and Spring~\cite{mehl2023spring}.
We compare against RigidMask (RM)~\cite{yang2021rigidmotion}, a SOTA two-frame approach.
It is trained on the SceneFlowDatasets~\cite{mayer2016scenflow}; however, they also train a version for driving scenarios (RM-Drive).
To evaluate, we report foreground IoU (higher is better).
\setlength{\columnsep}{2pt}
\setlength{\intextsep}{0pt}
\begin{wrapfigure}[3]{r}[5pt]{.41\columnwidth}
  \resizebox{0.4\columnwidth}{!}{
    \begin{tabular}{l|c|c}
      & VKITTI & Spring \\
    \hline
    RM & $32.4$ & $17.0$\\ 
    RM-Drive & $36.5$ & $8.4$\\ 
    L4P & $\mathbf{56.0}$ & $\mathbf{21.5}$ \\
    \end{tabular}
    }
\end{wrapfigure}
On both datasets, our video-based approach performs better.
Note that while fine-tuning on driving scenes helps RigidMask (RM-Drive) on VKITTI, it significantly hurts performance on Spring, highlighting the benefit of our model's generalization ability.
As shown in Figure~\ref{fig:vis_res} (third row), our approach performs better and can detect small motions. 
See more comparisons in the Supplementary.

\noindent\textbf{Camera pose estimation.}
We follow \cite{zhang2024raydiffusion} and represent a camera as a bundle of rays.
More specifically, each camera is represented by the $16\times16$ camera rays towards the image patch centers, 
and each ray by its 6-D Plücker coordinates. 
Our DPT dense head can be easily adapted to output a $16\times16$ grid of rays.
From the estimated ray bundles, we can then recover camera poses by solving least-square optimizations for camera centers, rotation matrices and even camera intrinsics~\cite{zhang2024raydiffusion}.
\setlength{\columnsep}{2pt}
\setlength{\intextsep}{0pt}
\begin{wrapfigure}[4]{r}[5pt]{.57\columnwidth}
  \resizebox{0.56\columnwidth}{!}{
    \begin{tabular}{l|ccc}
         & Sintel      & TUM-dyn.       & ScanNet        \\
         \hline
        DUSt3R~[CVPR24]              & $0.290$       & $0.140$          & $0.246$          \\
        Spann3R~[3DV25]              & $0.329$       & $0.056$          & $0.096$ \\
        CUT3R~[CVPR25]                & $0.213$       & $\mathbf{0.046}$    & $0.099$    \\
        L4P (head FT)                  & $0.162$ & $0.060$          & $0.100$         \\
        L4P (e2e FT)                  & $\mathbf{0.132}$ & $0.055$          & $\mathbf{0.080}$         \\
        \end{tabular}
    }
\end{wrapfigure}
For evaluation, we follow \cite{zhang2024monst3r} and show results on both dynamic (Sintel and TUM-dynamics) and static scenes (ScanNet).
We provide ATE (lower is better) and focus on feedforward approaches using general architectures.
We show two versions: one by only fine-tuning the camera head (head FT), and one by fine-tuning the entire model end-to-end (e2e FT) for all tasks, including camera pose estimation.
In either settings, our approach performs favorably with recent approaches like DUSt3R~\cite{wang2024dust3r}, Spann3R~\cite{wang2024spann3r}, CUT3R~\cite{wang2025cut3r}.
Once we have the camera pose estimation, we can visualize the camera trajectory, depth and 3D tracks in the same reference system (Figure~\ref{fig:track}(b)).
Additional metrics and methods are in the Supplementary.
\section{Limitations}\label{sec:limitations}
%! TEX Root = ../main.tex

Our main limitation is the low input resolution ($224 \times 224$), inherited from VideoMAE to leverage pretrained weights. 
This limits the use of high-resolution information useful for many applications.
We emphasize this is not a fundamental limitation, and could be addressed through high-resolution fine-tuning of VideoMAE (following DINOv2~\cite{oquab2023dinov2}) or by incorporating convex upsampling layers (common in opticalflow approaches).
A second limitation is weaker performance on single-image depth estimation (Table~\ref{tab:depth}, NYUv2), since L4P is trained only on videos. 
Incorporating image-only training into pre-training of VideoMAE and fine-tuning of L4P could help close this gap.

\section{Conclusions}\label{sec:conclusion}
%! TEX Root = ../main.tex
We present a unified framework to solve multiple low-level 4D vision perception tasks, both dense and sparse.
We achieve this by adopting a strong pre-trained video masked auto-encoder and design lightweight task heads to harness its representation power.
Our simple yet versatile designs for task heads allow for effortless and generalizable adaptation to multiple 4D vision perception tasks.
\section{Acknowledgements}
We would like to thank Jan Kautz for the continuous discussions and for reviewing an early draft of the paper, Zhiding Yu and Hongxu (Danny) Yin for the initial discussions on video models, and Yiqing Liang for help with the data. 

{
    \small
    \bibliographystyle{ieeenat_fullname}
    \bibliography{main}
}

%! TEX Root = ../main.tex
\clearpage
\section*{\huge Supplementary Material}
\setcounter{page}{1}
\setcounter{section}{0}
\setcounter{table}{0}

In this supplementary document, we provide comparisons for the inference time, and discuss additional implementation details regarding architecture, dataset and training.
We also provide additional results for 4D reconstruction and 3D tracking.
Please refer to supplementary video for a high-level overview and video results and comparisons.

\section{Inference time}
Our method solves multiple perception tasks at once and it is attractive for any application that requires a low-level, holistic understanding of the scene.
We show that its performance is competitive with, and sometimes better than methods designed for individual tasks.
Its computational time also makes it appealing, as it solves these tasks jointly, with efficiency comparable with that of single-task methods. 
For a video clip of size $16\times224\times224$, our inference runs in around 300ms on an NVIDIA A6000, which corresponds to $\sim$19ms/frame, or $\sim$28ms/frame if we consider the overhead due to overlapping windows in long videos.
This runtime includes all the tasks we discuss in the paper: depth, flow, tracking, motion segmentation, and pose estimation.
Table~\ref{tab:inferencetime} shows the runtime of the SOTA methods we compare against in the paper and confirms that our joint estimation is indeed efficient.

\section{Training datasets}
\label{sec:supp_data}
Our video encoder~\cite{wang2023videomaev2} has been pre-trained on 1.35M video clips across various data sources using masked auto-encoding.
To fine-tune our model, we use a limited number of synthetic datasets covering a varying range of annotations, and rely on the priors from the video encoder for generalization.

\noindent
\textbf{Kubric~\cite{greff2022kubric}.}
This synthetic dataset has multi-object interactions with many random objects.
We use it to generate annotations for depth, flow, motion-based segmentation, camera pose estimation and 2D/3D tracking.
Each video is 24 frames long, and we use a total of 15k videos from the movi-e and movi-f subsets of the data.
Kubric provides meta-data for object and camera trajectories, which could be used to generate 3D tracks.
We follow official guidelines and generate the annotations for 3D tracking by sampling around 8-12k tracks in each video.

\noindent
\textbf{PointOdyssey~\cite{zheng2023pointodyssey}.}
We use this synthetic dataset for depth, camera pose and 2D/3D tracking annotations.
The dataset consists of 159 videos, averaging 2k frames long, with human and object interactions under different lighting and atmospheric conditions.
We sample smaller video clips from the long videos to form our dataset.

\noindent
\textbf{DynamicReplica~\cite{karaev2023dynamicstereo}.}
We use this synthetic dataset for depth, flow, camera pose and 2D/3D tracking annotations.
The dataset consists of 524 videos with humans and animals in motion, and we sample smaller video clips to form our dataset.
Since this dataset has higher fps videos, we sample videos with strides of 1, 2 and 4.

\noindent
\textbf{TartanAir~\cite{tartanair2020iros}.}
Finally, to increase the scene-level data distribution we use TartanAir to generate annotations for flow, depth and camera pose.
The data is collected in photo-realistic simulation environments, with both indoor and outdoor scenes, in the presence of various light and weather conditions.
We sample smaller video clips from this dataset.

\begin{table}
    \centering
    \resizebox{.8\columnwidth}{!}{
      \begin{tabular}{l|ccc}
        \multicolumn{1}{c|}{} & Task & GPU & Per-frame time (ms) \\
        \hline
        \multicolumn{1}{l|}{RAFT~\cite{teed2020raft}} & Optical flow & A100 & 29 \\
        \multicolumn{1}{l|}{MemFlow~\cite{dong2024memflow}} & Optical flow & A100 & 48 \\
        \multicolumn{1}{l|}{DepthAnything~\cite{yang2024depthanything}} & Depth & A100 & 10 \\
        \multicolumn{1}{l|}{DepthCrafter~\cite{hu2024depthcrafter}} & Depth & A100 & 436 \\
        \multicolumn{1}{l|}{CUT3R (CVPR 2025)~\cite{wang2025cut3r}} & Depth and Pose & A100 & 60.3 \\
        \multicolumn{1}{l|}{RigidMask~\cite{yang2021rigidmotion}} & Motion Seg. & V100 & 260 \\
        \multicolumn{1}{l|}{SpaTracker (w/o depth)~\cite{spatracker}} & 3D Track & A6000 & 17 \\
        \multicolumn{1}{l|}{Ours} & All & A6000 & 19 \\
      \end{tabular}
    }
    \caption{\textbf{Inference time.} We compare our per-frame inference time with several task-specific approaches and show at least comparable speed-wise to methods specialized for each task.}
    \vspace{-1em}
    \label{tab:inferencetime}
\end{table} 

\section{Architecture}
Our video encoder~\cite{wang2023videomaev2} processes video-clips of size $16 \times 224\times224$.
It uses a patch-size of $2\times14\times14$, which results in $P=2048$ video tokens, and an embedding dimension of $C=1408$.
It has 40 encoder blocks, and we use output from blocks $[14, 21, 28, 36]$ for DPT heads for dense tasks, while our sparse-head uses features from the last block (block 39).
Feeding the sparse and dense heads with tokens from different blocks allows us to maintain the performance on dense tasks while we fine-tune our model to train the memory mechanism for the tracking tasks.
We adapt the DPT head~\cite{ranftl2021dpt}, originally designed to work with image tokens, to work with video tokens. 
We do this by adding appropriate rearrangement operation that maps three-dimensional video tokens to video feature maps and replace 2D-convolutions with 3D.

For the sparse head, we adapt two-way attention and mask decoding from SAM~\cite{kirillov2023SAM}.
We make appropriate modifications to make it work with video tokens and for our tracking task.
For an input point query, we use 3D positional encoding of dimension $C$ to generate a point query token.
To identify all our input/output tokens we add learnable embeddings of dimension $C$.
These input/output tokens interact with the input video tokens via a two-way attention mechanism.
Following SAM, we only keep two layers of two-way attention mechanism.
In the mask decoding stage, we add rearrangement operations to map video tokens of size $P \times C$ to video feature maps of size $T \times H \times W \times C/8$.
We replace the 2D convolutions in the mask-decoder of SAM with 3D convolutions.
The video feature maps are finally mapped to 2D heatmap, depth-map and visibility map by doing a dot-product with the corresponding output tokens processed via the two-stage attention mechanism, as shown in Figure 3 in the main paper.

To enable memory mechanism for online tracking using overlapping windows, we introduce a track-feature token $\mathcal{F}$ that is passed to subsequent windows as an additional prompt.
We do this by mapping the processed feature token $\mathcal{F}$ from the previous window via a linear layer with input and output dimensions of $C$ and then passing it as input to the next window. 
To provide even more cross-window information, we pass the video tokens decoded by the two-way attention stage of the current window to the next.
We achieve this by projecting the decoded video tokens in the overlapping region via a linear layer with input and output dimensions of $C$, and by adding them to the corresponding incoming video tokens to the two-way attention stage in the next window.
For the non-overlapping region we add a learnable mask token of dimension $C$.

\begin{figure}
  \centering
  \includegraphics[width=\columnwidth]{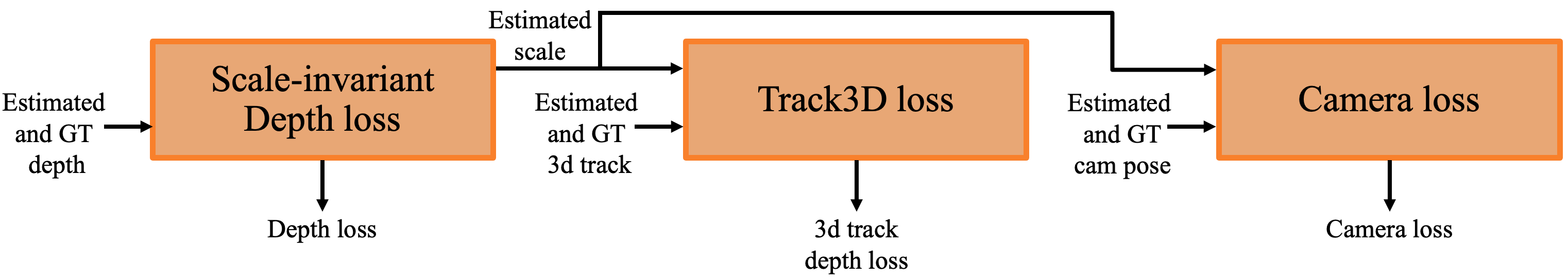}
  \caption{\textbf{Scale consistency.} To make the scales of different tasks consistent, we compute the scale factor from the depth loss and use it to constrain the scales for the 3D track depth and camera pose losses.}
  \vspace{-1em}
  \label{fig:scale_consistency}
\end{figure} 

%! TEX Root = ../../main.tex

\begin{figure*}[p]
    \centering
    \setlength{\tabcolsep}{2pt}
    \renewcommand{\arraystretch}{0}
    \begin{tabular}{cc}
        \includegraphics[width=0.39\textwidth]{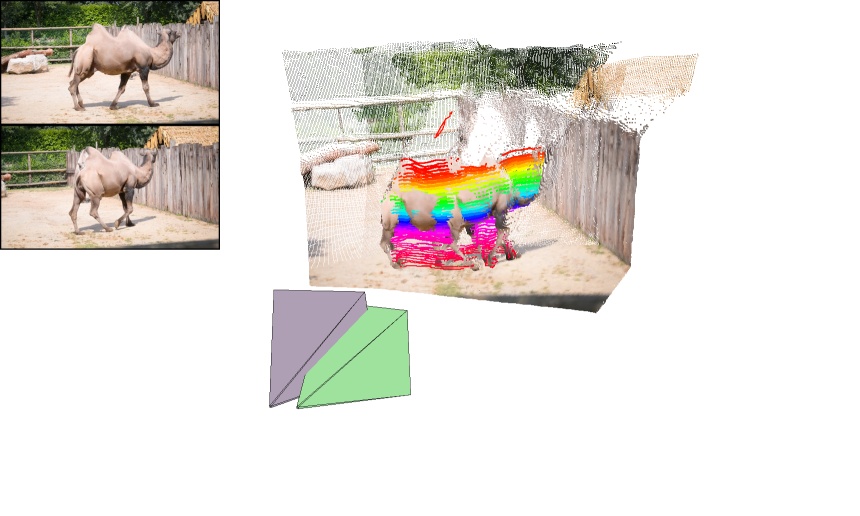} &
        \includegraphics[width=0.39\textwidth]{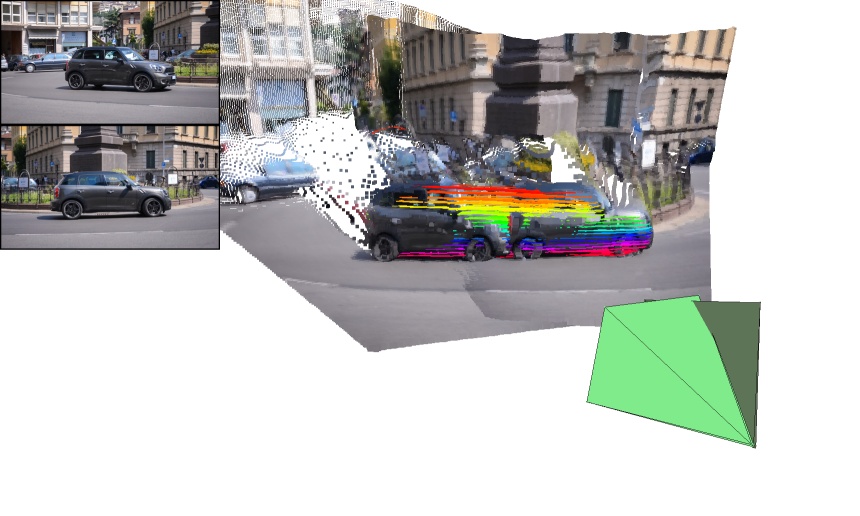} \\
        \includegraphics[width=0.39\textwidth]{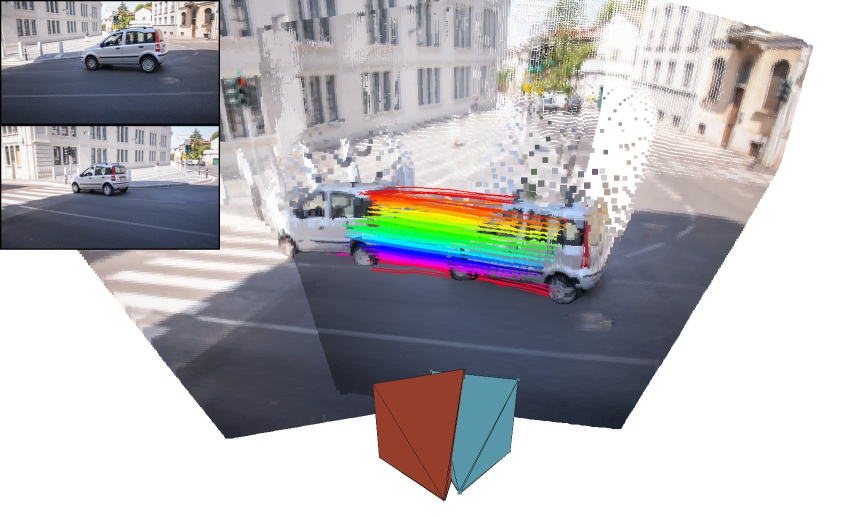} &
        \includegraphics[width=0.39\textwidth]{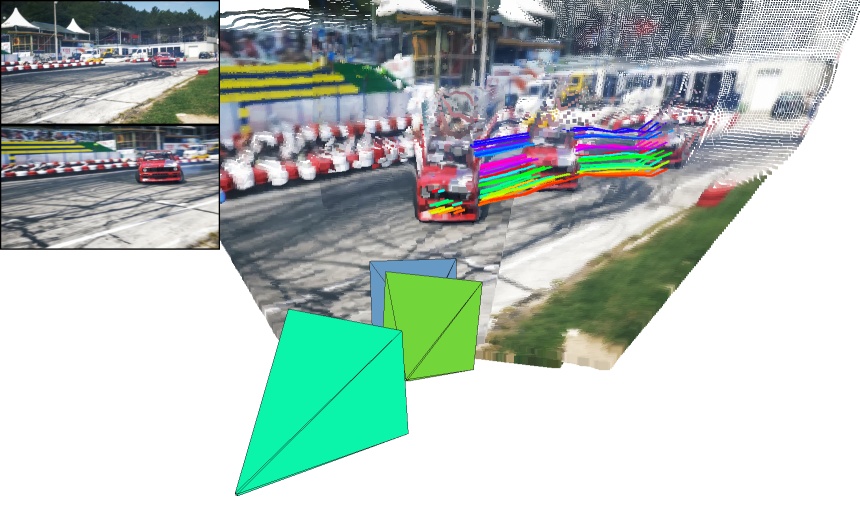}  \\
        \includegraphics[width=0.39\textwidth]{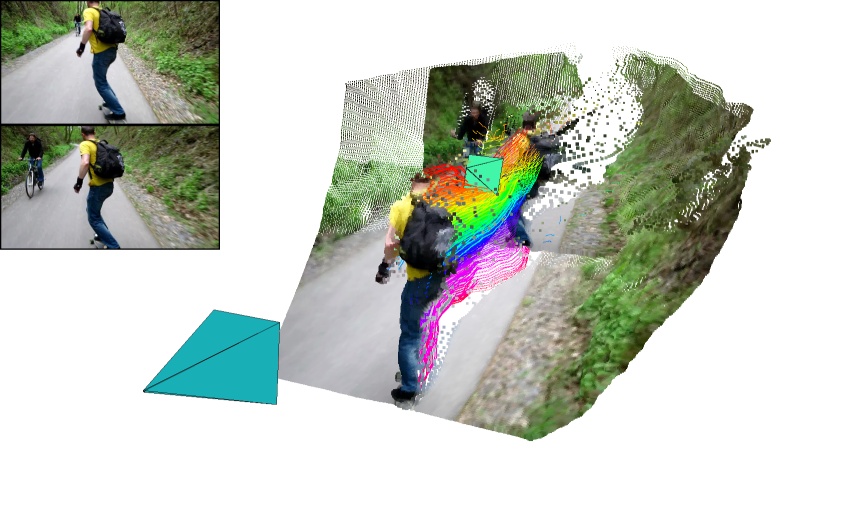} &
        \includegraphics[width=0.39\textwidth]{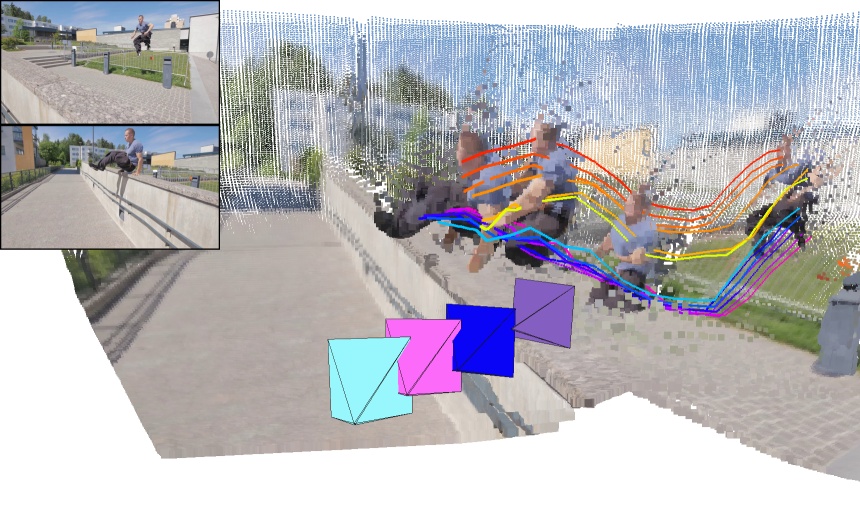}  \\
        \includegraphics[width=0.39\textwidth]{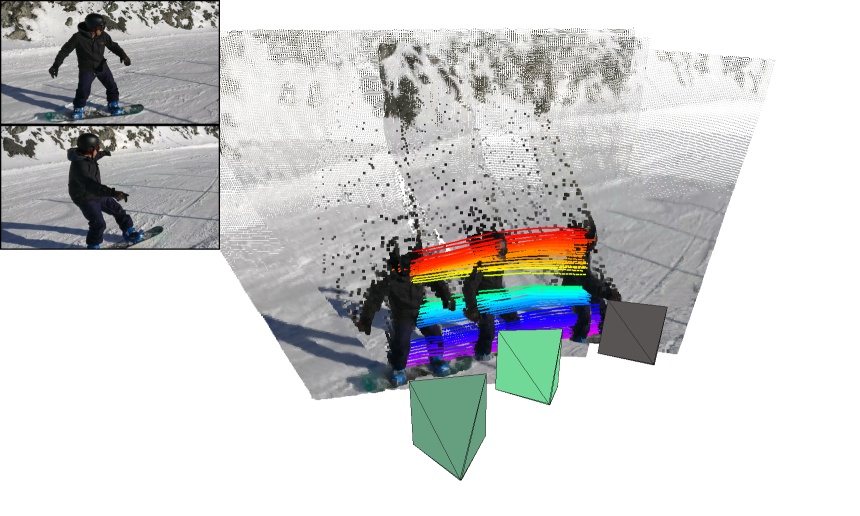} &
        \includegraphics[width=0.39\textwidth]{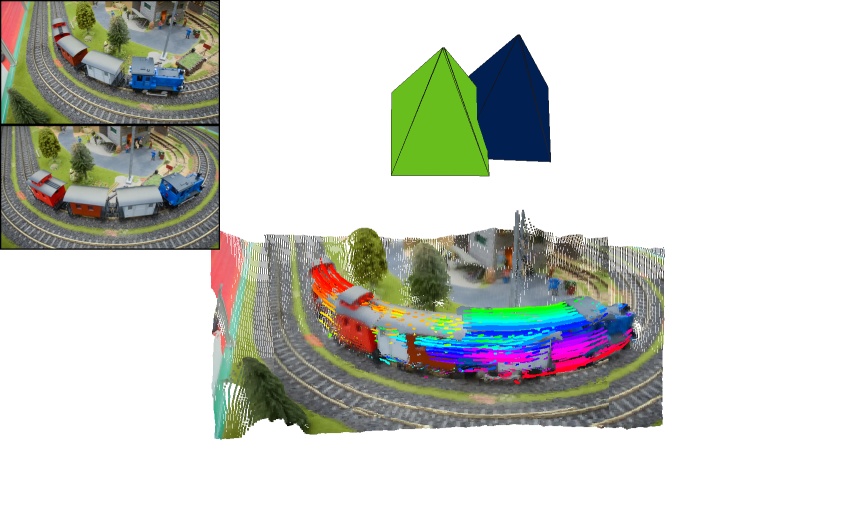} \\
        \includegraphics[width=0.39\textwidth]{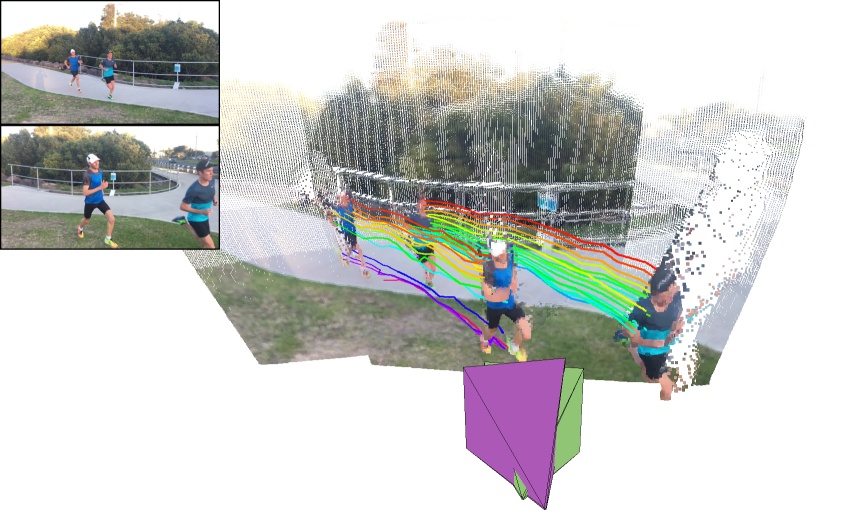} &
        \includegraphics[width=0.39\textwidth]{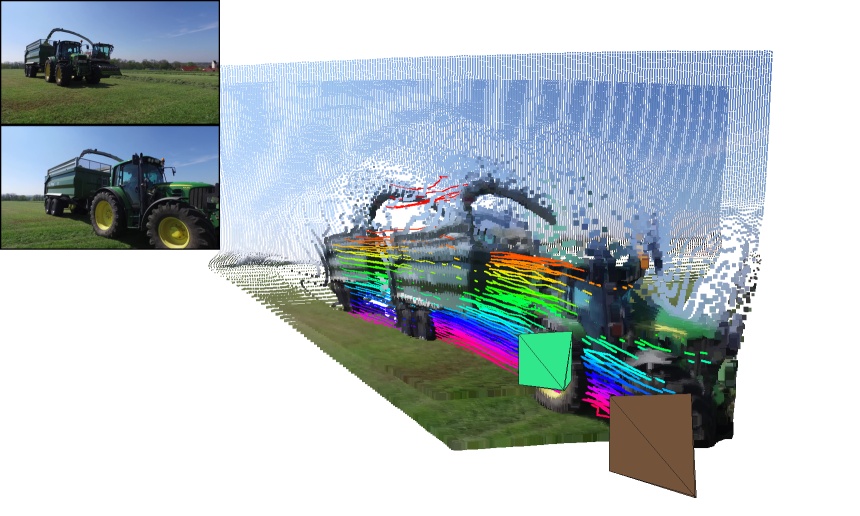} \\
    \end{tabular}
    \vspace{1em}
    \caption{
        \textbf{4D Reconstruction Results.} 
        We show example visualizations on diverse real-world sequences.
        For each scene, we jointly display the estimated depth maps, camera trajectories, and 3D point tracks, all aligned in a common reference frame.
        We also show the first and last input RGB frames in the top left corner for each example.
        Our approach enables efficient feedforward estimation of consistent and coherent 4D reconstructions across challenging dynamic environments.
    }
    \label{fig:4d_recon}
\end{figure*}

\section{Training}

We initialize our video encoder using a pretrained VideoMAE~\cite{wang2023videomaev2} and fine-tune our model in three stages.
All the stages use a batch size of 8 and are trained on a single 8-GPU (NVIDIA A100) node.
In the first stage we train all the parameters of our model for depth, flow and 3D tracking on a single window of $T=16$ frames and train only on the Kubric dataset.
In the second stage of training, we add the remaining three datasets and further fine-tune the model on a single window for all the tasks.
In this stage, since not all the datasets include all the tasks (flow missing in PointOdyssey and point tracking missing in TartanAir), we formed our batch by sampling two videos from each of the four datasets.
This ensures a fixed batch-size per-task across training iterations.
In the first two stages, for each video we estimate all $T$ frames for dense tasks and for 3D tracking we construct a batch of 80 tracks and select point queries randomly across the visible parts of tracks.

In the third stage, we further fine-tune our model for the tracking tasks using unrolled-window training and the memory mechanism for online tracking.
We train on videos of length $S=40$, windows of size 16 frames and a stride of 8, which results in 4 overlapping windows, and we construct a batch of 48 tracks during this stage, but generate point queries only in the first 20 frames to force the network to learn long-range tracking.
To train this stage efficiently, we only train for the 2D and 3D tracking tasks and freeze all the parameters, except the last three layers (37-39) of the video encoder and the sparse head.
Note that since other dense tasks use the output of layers smaller than 37, their performance remains unaffected by this stage.
For all the stages we use AdamW optimizer with a maximum learning rate of $5\times10^{-5}$, beta values of $(0.9, 0.999)$, weight-decay value of $0.05$, gradient-clipping value of $0.05$ and trained using mixed-precision.
We use cosine annealing strategy for learning rate where we use first 5k iterations to increase the learning rate to the maximum value.
Training takes around four days for all the stages, one day each for the first two stages and two days for the third stage.

\noindent\textbf{Losses.}
For depth, we use SILog~\cite{eigen2014depth} loss, for flow, we use L1-loss on the estimated uv-offsets, and for tracking, we use L1-loss for the 2D track positions, scale-invariant loss for the track depth (similar to dense depth), and binary-cross entropy loss for the track visibility.
For the additional tasks of motion-based segmentation and camera pose estimation, we use binary cross-entropy loss and L1 loss, respectively.
We found the loss weights empirically by first weighting the losses to be in the same order of magnitude and then doing a small hyperparameter search around those weights.
We use the loss weights of $20$ for the flow and depth losses. For tracking, we use loss weights of $1.0$, $20.0$, and $15.0$ for 2D track position loss, track depth loss and track visibility loss, respectively.

The scale-invariant loss~\cite{eigen2014depth} we use for depth and 3D track depth is computed as shown below:
\begin{equation}
    L(y,y^{*}) = \frac{1}{N} \sum_{i=1}^{N} \left( \log(y_i) - \log(y_i^*) + \alpha(y,y^{*}) \right)^2
    \label{eq:silog_loss}
\end{equation}
\begin{equation}
    \alpha(y,y^{*}) = \frac{1}{N} \sum_{i=1}^{N} \left( \log(y^{*}) - \log(y) \right)
    \label{eq:silog_scale}
\end{equation}
where $y$ and $y^{*}$ are the predicted and ground-truth entities, respectively.
For any prediction $y$, $e^{\alpha(y,y^{*})}$ is the scale that best aligns it to the ground truth.

\noindent\textbf{Scale consistency.}
While we solve depth, 3D tracking and camera pose estimation within a single unified framework, they could have different scale factors if proper care is not taken.
We need additional constraints in our loss functions to make them consistent with one another.
To achieve this, we follow a simple approach and constrain the scale for the 3D track depth and camera pose losses to be the same as that of the depth loss. 
This is done by computing losses in a specific order as shown in Figure~\ref{fig:scale_consistency}.
First, we compute the scale-invariant loss and the scale factor $\alpha$ for the depth estimation following Equations~\ref{eq:silog_loss} and~\ref{eq:silog_scale}.
We then use the estimated scale factor from depth loss to constrain the scales for the 3D track depth loss and camera pose loss.
This allows us to constrain the scales of different tasks and to visualize them together in a common reference frame, as shown in the 4D reconstruction visualizations in Figure~\ref{fig:4d_recon}.

\section{Motion-based segmentation}
For motion-based segmentation, we generate GT annotations for training using Kubric dataset.
We use the camera-pose and the 3D track information to detect which 3D tracks come from dynamic vs. static objects.
We combine this information with the provided rigid-object segmentations to generate annotations for motion-based segmentation.
For evaluation datasets, we use Virtual KITTI~\cite{cabon2020vkitti2} and Spring~\cite{mehl2023spring} datasets.
Both datasets provide camera poses, depth and scene-flow information.
We combine depth and camera information to estimate scene-flow due to the camera-motion.
This allows us to compute scene-flow due to dynamic objects and generate ground-truth annotations for motion-based segmentation.
We train the DPT head for this task using a binary cross-entropy loss.
We use a batch size of 8, use random affine augmentations, and train for 25k iterations.

\section{Camera pose estimation}
%! TEX Root = ../main.tex

Here we provide additional details and results for the pose estimation task.

%! TEX Root = ../main.tex
\begin{table}
\begin{center}
    \resizebox{\columnwidth}{!}{
    \begin{tabular}{lccccccccc}
        & \multicolumn{3}{c}{Sintel ($\sim$50 frames)}   & \multicolumn{3}{c}{TUM-dynamics (90 frames)} & \multicolumn{3}{c}{ScanNet (90 frames)} \\
        \cmidrule(lr{0.1em}){2-4}\cmidrule(lr{0.1em}){5-7}\cmidrule(lr{0.1em}){8-10}
Method            & ATE $\downarrow$    & RPE-trans $\downarrow$ & RPE-rot $\downarrow$ & ATE $\downarrow$      & RPE-trans $\downarrow$  & RPE-rot $\downarrow$  & ATE $\downarrow$       & RPE-trans $\downarrow$    & RPE-rot $\downarrow$   \\
\midrule
        DROID-SLAM$^{*\dagger}~\cite{teed2021droid}$ & $0.175$  & $0.084$ & $1.912$   & $\mathbf{0.021}$ & $\mathbf{0.013}$ & $\mathbf{0.358}$   & $\mathbf{0.046}$ & $\mathbf{0.015}$ & $\mathbf{0.449}$ \\
        DUSt3R~\cite{wang2024dust3r}$^*$       & $0.417$  & $0.250$     & $5.796$   & $0.083$    & $0.017$      & $3.567$    & $0.081$     & $0.028$        & $0.784$     \\
        MonST3R~\cite{zhang2024monst3r}$^*$    & $\mathbf{0.111}$  & $\mathbf{0.044}$     & $\mathbf{0.869}$   & $0.098$    & $0.019$      & $0.935$    & $0.077$     & $0.018$        & $0.529$     \\
        \midrule
        DUSt3R~\cite{wang2024dust3r} (w/o GA)  & $0.290$  & $0.132$     & $7.869$   & $0.140$    & $0.106$      & $3.286$   & $0.246$     & $0.108$        & $8.210$     \\
        Spann3R~\cite{wang2024spann3r}         & $0.329$  & $0.110$     & $4.471$   & $0.056$    & $0.021$      & $0.591$    & $0.096$     & $0.023$        & $0.661$     \\
        CUT3R~\cite{wang2025cut3r}             & $0.213$  & $0.066$     & $0.621$   & $\mathbf{0.046}$ & $\mathbf{0.015}$ & $0.473$    & $0.099$ & $\mathbf{0.022}$ & $0.600$  \\
        
        \methodName (head FT)           & $0.162$ & $0.072$ & $0.999$    & $0.060$ & $0.019$ & $0.935$   & $0.100$ & $0.031$ & $0.940$    \\ 
        \methodName (head FT)$^\dagger$ & $0.149$ & $0.072$ & $0.417$    & $0.062$ & $0.017$ & $0.552$   & $0.095$ & $0.029$ & $0.584$    \\ 
        \methodName (e2e FT)            & $0.132$ & $0.056$ & $0.577$    & $0.055$ & $0.019$ & $0.675$   & $0.080$ & $0.028$ & $0.705$    \\ 
        \methodName (e2e FT)$^\dagger$  & $\mathbf{0.115}$ & $\mathbf{0.049}$ & $\mathbf{0.379}$    & $0.048$ & $\mathbf{0.015}$ & $\mathbf{0.462}$   & $\mathbf{0.075}$ & $0.025$ & $\mathbf{0.477}$    \\ 
        
\bottomrule
\end{tabular}
    }
\end{center}
\caption{\textbf{Additional camera pose estimation results.} 
We consider two types of approaches: optimization-based and feedforward approaches. 
``DUSt3R (w/o GA)" is an online variant of DUSt3R, which aligns all video frames with first frame, without using global alignment (GA).
$\dagger$ indicates methods that use camera intrinsics at inference time, and $*$ indicates the use of global optimization.
Our \plucker-based approach works with or without camera intrinsics.
We show results for two versions.
The ``head FT" version only fine-tunes the camera head while keeping the L4P video encoder frozen. 
The ``e2e FT" version fine-tunes the entire model end-to-end for all tasks (depth, flow, 2D/3D tracking and camera pose).
}\label{tab:pose}
\vspace{-1em}
\end{table}

\noindent\textbf{Architecture.} 
We modify the DPT head used in the previous dense tasks to output $16\times16$ instead of full dense resolution, 
matching the resolution suggested in \cite{zhang2024raydiffusion}. 
This is achieved by changing the sampling rates at several layers within the DPT head,
The last layer produces 6 channels, representing the 6-D \plucker coordinates $(d, o\times d)$, where $d$ is the ray direction and $o$ is a point on the ray.

\noindent\textbf{Training.}
We freeze the video encoder and train the camera pose estimation head with all datasets described in Section~\ref{sec:supp_data}.
GT rays are constructed for all samples using provided camera intrinsics and poses. 
We scale the pose using the scale factor estimated from the depth loss to make the pose and depth scales consistent with each other.
We apply L1-loss on the \plucker coordinates, and use 16-frame clips for training.
We use a batch size of 8 and train for 200k iterations.
We also train a version of the model that is trained end-to-end for depth, flow, 2D and 3D tracking, and camera pose estimation.
We report the numbers for both head-only fine-tuning and end-to-end fine-tuning in Table~\ref{tab:pose}.

\noindent\textbf{Testing.} 
At inference time, output \plucker coordinates can be converted to traditional camera pose representations for evaluation by solving least-square optimizations, as done in~\cite{zhang2024raydiffusion}.
We consider and evaluate under two settings, one where camera intrinsics are provided at inference time, and one where they are not. 
The evaluation datasets, Sintel, TUM-dynamics, and ScanNet, all contain sequences longer than 16 frames, and therefore require inference beyond a single $T=16$ window.
To handle sequences longer than 16 frames, we perform inference on overlapping 16-frame windows with a stride of 8. 
We then align the predictions between consecutive windows using a 3D similarity transformation. 
This transformation is computed using the estimated depths and camera poses from the overlapping frames of adjacent windows, leveraging the fact that our depth and camera pose predictions are trained to have a consistent scale.
Specifically, we lift the depth maps to 3D point clouds using the estimated camera poses within each window, and then estimate the 3D similarity transform to align the camera poses across windows. We compare our results with both optimization-based and feedforward approaches, as shown in Table~\ref{tab:pose}.

\section{Additional results}

\noindent\textbf{4D reconstruction.}
We show additional visualization results for 4D reconstruction, where we jointly visualize the estimated depth maps and 3D point tracks aligned in a common reference frame using the estimated camera poses and camera intrinsics.
These results are generated using our efficient feedforward approach.
We show additional video results for 4D reconstruction in our supplementary video.

\noindent\textbf{Additional 3D tracking results.}
Evaluating 3D tracking performance on the TapVid3D~\cite{koppula2024tapvid3d} benchmark is computationally intensive, given the large number of videos and annotations in the dataset. 
Therefore, in addition to the full evaluation results presented in the main paper, we also report our results on the minival subset, which contains 150 videos.
The 2D AJ, 2D APD, 3D AJ, 3D APD and OA metrics for our approach are $49.0$, $65.0$, $10.9$, $17.5$ and $87.6$ respectively.
We can compare these numbers with the minival results reported for Seurat~\cite{cho2025seurat}, a recent 3D tracking method that utilizes 2D point tracks from the state-of-the-art CoTracker~\cite{karaev2023cotracker} and depth maps from DepthPro~\cite{bochkovskii2024depthpro}.
The 3D AJ and 3D APD numbers reported by authors are $11.4$ and $18.1$, which are slightly better than ours.

\end{document}